\documentclass[11pt,a4paper]{article}
\usepackage[a4paper,left=1.6cm,right=1.6cm,top=1.6cm,bottom=1.8cm,footskip=0.8cm]{geometry}
\usepackage[T1]{fontenc}
\usepackage{mathptmx}
\usepackage{amsmath,amssymb}
\usepackage{graphicx}
\usepackage{booktabs}
\usepackage{array}
\usepackage{xcolor}
\usepackage{pifont}
\usepackage[numbers,sort&compress]{natbib}
\usepackage[hidelinks]{hyperref}
\usepackage{xurl}
\usepackage{enumitem}
\usepackage{titlesec}
\usepackage{caption}
\usepackage{longtable}
\usepackage{multicol}
\usepackage{needspace}

\titleformat{\section}{\normalfont\large\bfseries}{\thesection}{0.6em}{}
\titleformat{\subsection}{\normalfont\normalsize\bfseries}{\thesubsection}{0.6em}{}
\titlespacing*{\section}{0pt}{14pt plus 3pt minus 2pt}{2pt plus 1pt}
\titlespacing*{\subsection}{0pt}{10pt plus 2pt minus 1pt}{0pt plus 1pt}
\setlist{itemsep=3pt,topsep=4pt,parsep=0pt,partopsep=0pt}

\AtBeginDocument{\postdisplaypenalty=10000\predisplaypenalty=10000}
\AtBeginDocument{\setlength{\abovedisplayskip}{10pt plus 2pt minus 2pt}\setlength{\belowdisplayskip}{8pt plus 2pt minus 2pt}}

\newcommand{\figcap}[2]{\caption[#1]{\textbf{#1}\newline #2}}
\newcommand{\tabnote}[1]{\par\vspace{5pt}\begin{minipage}{0.9\linewidth}\centering\footnotesize #1\end{minipage}\par}

\makeatletter
\def\tagform@#1{\maketag@@@{Equation~(\ignorespaces#1\unskip\@@italiccorr)}}
\makeatother
\newenvironment{where}{\par\nopagebreak\vspace{2pt}\noindent\begin{minipage}{\linewidth}Where:\par\vspace{1pt}\noindent
  \begin{tabular}{@{\hspace{2em}}p{0.16\linewidth}p{0.6\linewidth}l@{}}}{\end{tabular}\end{minipage}\par\vspace{9pt}}

\newcommand{\yes}{\ding{51}}
\newcommand{\no}{\textcolor{black!35}{\ding{55}}}
\newcommand{\prt}{$\circ$}
\graphicspath{{fig/}}

\begin{document}

\begin{center}
{\LARGE\bfseries Predict the Swing, Match the Speed: Robotic Pipe Stabbing\\[2pt] With Learned Stereo Vision and Residual Reinforcement Learning\par}
\vspace{10pt}
{\large Mohamed Ashraf Mostafa\textsuperscript{1,*}, Taher El-Fakharany\textsuperscript{1,2}, Samir Khaled\textsuperscript{1,2}\par}
\vspace{6pt}
{\small\itshape
\textsuperscript{1}Mining and Petroleum Engineering Department, Faculty of Engineering, Al-Azhar University, Cairo, Egypt\\
\textsuperscript{2}Petroleum Department, Faculty of Engineering and Technology, Future University in Egypt (FUE), Cairo, Egypt\par}
\vspace{4pt}
{\small \textsuperscript{*}Corresponding author. Email address: \texttt{mohamedashraf19116@azhar.edu.eg} (Mohamed Ashraf Mostafa)\par}
\end{center}
\vspace{2pt}
\noindent\rule{\linewidth}{0.4pt}
\par\noindent\textbf{Abstract}\par\vspace{2pt}
\noindent A joint of pipe run into the hole swings from the elevator over the well and strikes a fixed stabbing guide at whatever speed the swing has. StabBot, a robotic stabbing system, predicts the swing and meets the pipe at its own speed. A pendulum Kalman filter tuned from the pipe tally tracks the joint in stereo images. A six-axis arm with an open V-shaped roller head closes in at matched velocity, confirms the last millimetres by touch, aims the pin over the box when the elevator hangs off centre, and follows the block down with lateral compliance. A ResNet-18 keypoint network locates the pipe, pin nose and box mouth in 3-D, and a residual reinforcement-learning policy refines the capture. On 200 paired random jobs with nine sizes from 2-3/8-in. tubing to 20-in. casing, StabBot seated every joint with the pipe touching only rollers, against 74\% for a conventional fixed guide. Median contact speed fell from 18.4 to 2.3\,cm/s and median peak head-pipe force from 11.3 to 1.4\,kN, and the force was lower in all 200 jobs (Wilcoxon $p<10^{-30}$). The network locates the pipe with 0.4-mm median error in 2.3\,ms per stereo pair. With the network in the loop, success stayed at 100\% under 20\,mm of added noise, 300-ms delay, 85\% frame loss, 20-mm calibration offset, 20-cm elevator offset and 40-cm swings. All results come from a 1:4 Froude-scaled MuJoCo model and are reported at full scale.

\vspace{4pt}\noindent\textbf{Keywords:} Drill floor automation; Pipe handling; Robotic stabbing; Kalman filter; Keypoint detection; Residual reinforcement learning
\par\noindent\rule{\linewidth}{0.4pt}

\begin{figure}[!htbp]
\centering
\includegraphics[width=0.497\textwidth]{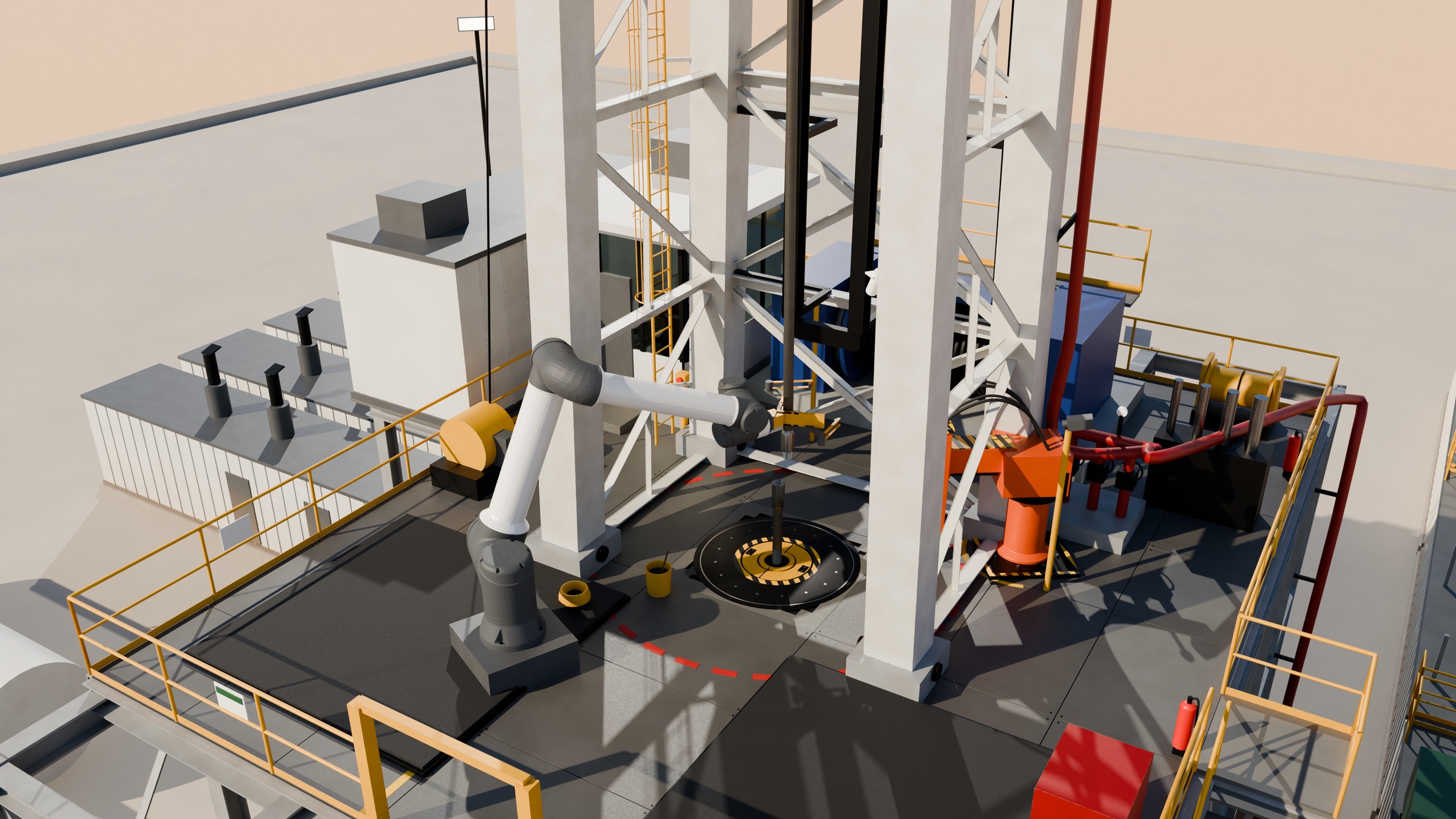}\hfill
\includegraphics[width=0.497\textwidth]{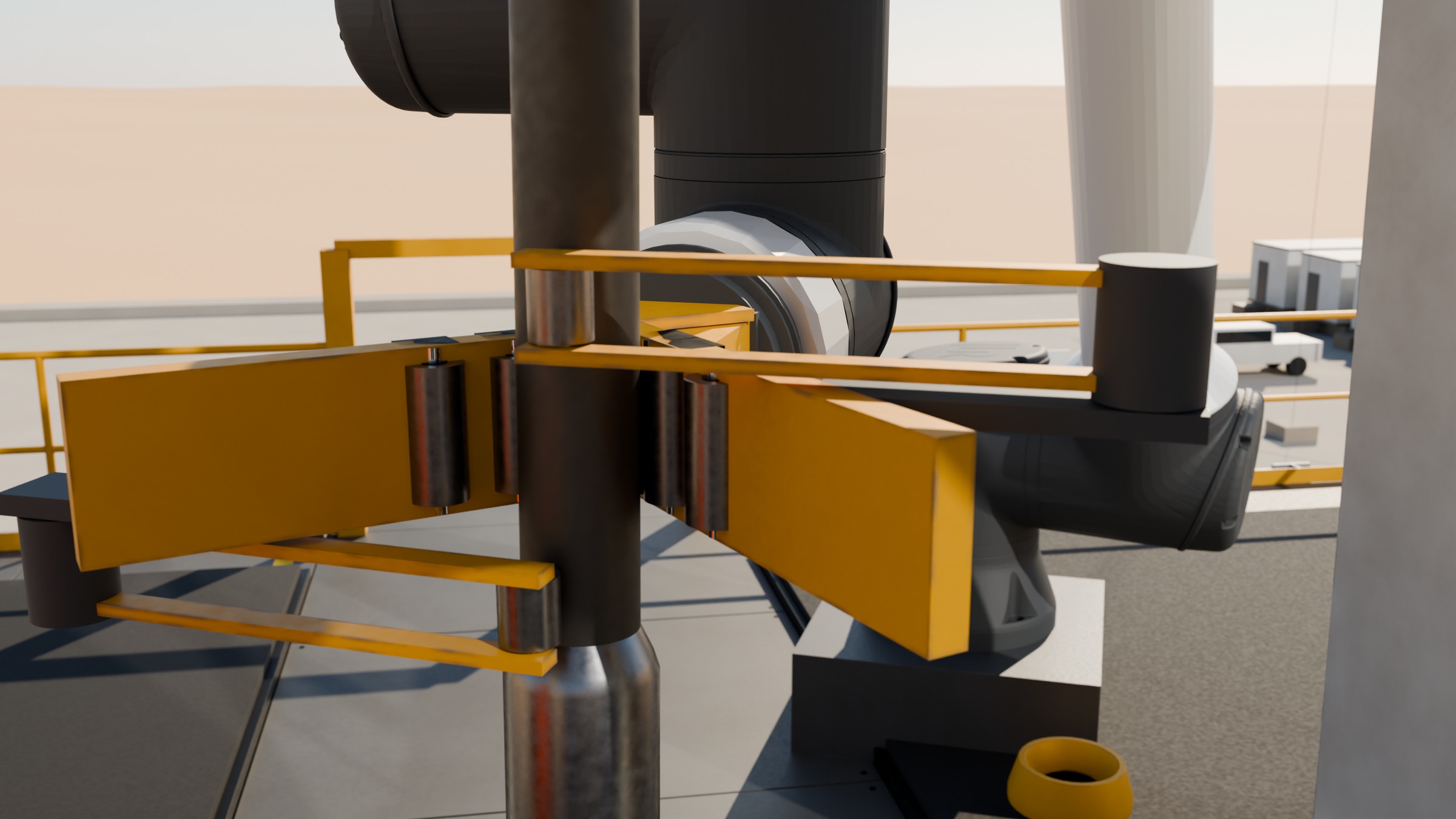}
\figcap{StabBot on a land-rig drill floor}{Left: the arm stands beside the V-door lane, the joint hangs from the elevator over the well and the box is held in the slips. Right: the open V head has met the swinging 5-in. drillpipe at its own speed and closed its two rotary clamps; the pipe rests on free rollers only. Every frame is a replay of the recorded physics, rendered at full scale.}
\label{fig:teaser}
\end{figure}

\section{Introduction}
A joint of pipe hanging from the elevator is a pendulum about 9.6\,m long and up to 1.3\,t heavy. Before it can be made up, its pin has to be brought to rest over a box held in the rotary, a target a few centimetres wide. On rigs without a stabbing robot this is done by a floorhand with a hand on the pipe, sometimes helped by a tagline or a passive stabbing guide. The moving pipe is a known hazard: in a U.S. Bureau of Safety and Environmental Enforcement report, the bottom of a stand being stabbed whipped when the link tilt moved and fractured a floorhand's fingertip \citep{bsee2018energyxxi}. It also wears equipment. A pin that strikes the box face can damage threads and seals, and a guide that the pipe hits at full swing speed takes a shock load.

Drill floor robots have been field-tested and are offered commercially \citep{raunholt2021field,nabors2025rzr}, and full robotic drill floors have been studied for land rigs and drillships \citep{cappuccio2019full}. Stabbing guides have been fitted to drill floor robots and robotic roughnecks \citep{nabors2025rzr,mikalsen2024stabbing}, and a remotely operated stabber has been used in completion operations \citep{mwansa2021enhancing}. In the systems we reviewed, the tool waits at well centre until the swinging pipe reaches it, so the first contact takes place at whatever speed the swing has at that instant. We are not aware of a published system that estimates the swing of the joint and plans the capture around it (Section~\ref{sec:related}).

StabBot instead watches the swing for 0.8\,s, fits a pendulum model whose frequency comes from the pipe tally, shadows the predicted pipe and closes in along a minimum-jerk path \citep{flash1985coordination} so that the guide head arrives moving with the pipe. The face rollers then sense contact, and the last millimetres are closed by touch. Once the pipe is held, the robot brings it to rest over the box. When the elevator hangs off centre, the hanging pipe is tilted, and the robot places the pin nose over the box instead of the pipe axis at head height. As the block slacks off, wrist admittance lets the taper finish the alignment.

The study addresses two questions. The first is whether a robot that knows where the pipe will be can stab it more reliably and more gently than one that waits. The second is how much each of the tools usually proposed for such a robot (a predictive controller, a learned perception model and a learned control correction) adds. Both are answered with one paired benchmark in which every method runs on the same 200 random jobs. The contributions of this work are as follows.
\begin{enumerate}
\item \emph{Swing-aware capture} for hanging tubulars: a tally-tuned pendulum Kalman filter, velocity-matched capture, a force-confirmed final approach and tilt compensation for an off-centre elevator (Section~\ref{sec:method}).
\item \emph{One open guide head for nine sizes}, 2$\frac{3}{8}$-in. tubing to 20-in. casing, on which the pipe touches only free rollers, with at least 20\,mm clearance on every other part (Section~\ref{sec:system}).
\item \emph{A stereo keypoint network} that locates the pipe, the pin nose and the box mouth in 3-D with sub-millimetre median error and runs in the control loop at 30\,Hz.
\item \emph{A residual reinforcement-learning policy} that lowers peak force by a further 22\% relative to the model-based controller it corrects. A reward failure met during its development is reported with its fix.
\item \emph{A 200-job paired benchmark} with robustness, unseen-condition and ablation tests. StabBot seats 100\% of joints against 74\% for a fixed guide, and its median contact speed and median peak force are both lower by a factor of 7.8 (Section~\ref{sec:results}).
\end{enumerate}
Code, trained models, recorded jobs, drawings and videos will be released on acceptance.

\section{Related Work}\label{sec:related}

\subsection{Drill floor automation and stabbing}
Robotics in oil and gas has been reviewed for onshore and offshore operations \citep{shukla2016onshore,shukla2016offshore}, and full robotic drill floors have been studied for land rigs and drillships \citep{cappuccio2019full}. Stabbing itself is already mechanised in several forms. In TotalEnergies' tests of Canrig drill floor and pipe handling robots, a stabbing device was added to the dual-arm pipe handler so that stabbing succeeds even with slightly bent pipe \citep{raunholt2021field}. ADNOC Onshore and Frank's International deployed a remotely operated stabber for completion tubulars smaller than 4.5\,in. \citep{mwansa2021enhancing}. The Nabors RZR arm carries a stabbing guide at well centre \citep{nabors2025rzr}, and patents describe a stabbing guide on a robotic roughneck \citep{mikalsen2024stabbing}, passive spring-loaded funnels that centre the pin on the box \citep{saunders2023passive}, and drill floor robots with stabbing-guide and funnel end effectors \citep{holck2022robotic}. The closest industrial work treats tripping as pendulum-like: the robot holds the bottom of the stand and keeps its gripper axis pointed at the elevator pivot along the arc \citep{moon2022nov}. It guides the stand along a planned path and does not estimate the free swing or intercept it. Vision on the drill floor measures stick-up height \citep{vintervold2017tripping}, and a patented system estimates stick-up height, position and orientation from CNN recognition of pipe and tool joint combined with time-of-flight and lidar data \citep{blakely2023stickup}, and visual servoing lets a pipe manipulator grasp a quasi-static pipe \citep{zhang2019drillpipe}. None of these sources describes estimating or predicting the swing of a hanging joint.

\subsection{Swinging loads, catching and fine motion}
Crane control damps the swing of a suspended payload from the hoist side \citep{ramli2017crane}, and vision has been used to estimate the pendulum's cable length online for a knuckle boom crane \citep{tysse2021vision}. The closest robotic precedent grips a crane-hung steel beam: Jung \emph{et al.}~\citep{jung2007gripping} model the beam as a pendulum and, assuming the instant of impact is known in advance (for example from a camera), apply a pre-acting force to absorb the shock. Catching research has long tracked and grasped moving targets \citep{allen1993tracking,kim2014catching}, and soft catching moves with a flying object before closing on it, which lowers the impact \citep{salehian2016soft,zhao2023impact}. StabBot adapts that velocity matching to a pendulum-suspended tubular and follows the catch with an insertion. Guarded moves, in which a robot advances slowly until it feels contact, are a classic tool for fine motion \citep{lozanoperez1984automatic}.

\subsection{Keypoint vision}
Heatmap keypoint networks \citep{xiao2018simple} with a differentiable soft-argmax readout \citep{sun2018integral} reach sub-pixel accuracy; later work improves the decoding \citep{zhang2020dark} and the backbone and head \citep{sun2019hrnet,xu2022vitpose,jiang2023rtmpose}. Single-stage detectors that regress keypoints, such as YOLO-Pose \citep{maji2022yolopose}, are judged at loose tolerances. Keypoints have also proved a compact state for manipulation \citep{manuelli2022kpam}, and 3-D keypoints can be estimated directly from stereo pairs \citep{liu2020keypose}. The published vision work on drill floors finds pipe with bounding boxes or measures stick-up, and none of the studies we reviewed locates the pipe, the pin and the box as 3-D landmarks.

\subsection{Residual reinforcement learning}
Learning a correction on top of a conventional controller \citep{johannink2019residual,silver2018residual} keeps the base controller's behaviour and lets the policy improve on it, and has been applied to industrial insertion \citep{schoettler2020deep}. Learned insertion with force and vision \citep{inoue2017deep,lee2019making,luo2019variable} shows the value of contact sensing in the last millimetres, on static targets. Two pitfalls are well documented: unsmoothed actions excite hardware \citep{mysore2021caps}, and a policy exploits any gap between reward and objective \citep{krakovna2020specification}. Both arose in this work, and the remedy is described with the design lessons in Section~\ref{sec:results}.

Table~\ref{tab:related} sets the closest work against StabBot. Pendulum models of suspended loads, velocity-matched catching and learned pipe vision have each been studied separately. We found no published system that estimates the free swing of a hanging joint online and intercepts it at matched velocity for stabbing, that locates pipe, pin and box with a learned 3-D model on a moving joint, or that compares such a system with a conventional guide on a common paired benchmark. The two nearest systems differ in kind. Jung \emph{et al.}~\citep{jung2007gripping} assume a known instant of impact and apply a pre-acting force to absorb it, without matching the velocity of the beam or inserting it, and the pivot-pointing robot of Moon \emph{et al.}~\citep{moon2022nov} guides a stand it already holds along a planned arc, without estimating a free swing.

\begin{table}[!htbp]
\centering
\caption{StabBot against the closest published work}
\label{tab:related}
\small
\begin{tabular}{@{}l c c c c c@{}}
\toprule
Work & M & S & V & T & L\\
\midrule
Floorhand + tagline & \yes & \no & \prt & \yes & \no\\
Passive funnel \citep{saunders2023passive} & \no & \no & \no & \no & \no\\
Robot + stabbing guide \citep{raunholt2021field,nabors2025rzr,mikalsen2024stabbing} & \no & \no & \no & \prt & \no\\
Remote stabber \citep{mwansa2021enhancing} & \no & \no & \no & \no & \no\\
Pivot-pointing tailing \citep{moon2022nov} & \yes & \prt & \no & \no & \no\\
Stick-up vision \citep{vintervold2017tripping,blakely2023stickup} & \no & \no & \no & \no & \prt\\
Vision pipe grasp \citep{zhang2019drillpipe} & \no & \no & \no & \no & \prt\\
Crane beam gripping \citep{jung2007gripping} & \yes & \yes & \no & \no & \no\\
Soft catching \citep{salehian2016soft,zhao2023impact} & \yes & \no & \yes & \no & \no\\
\textbf{StabBot (this work)} & \yes & \yes & \yes & \yes & \yes\\
\bottomrule
\end{tabular}
\tabnote{M: moving (swinging) pipe; S: swing model; V: capture at matched velocity; T: touch-confirmed approach; L: learned 3-D pipe, pin and box landmarks. \yes\ addressed; \prt\ partly; \no\ not addressed in the published description.}
\end{table}

\section{System}\label{sec:system}

\subsection{Rig model and scaling}
The drill floor, the hanging joint, the robot and the guide head are modelled in MuJoCo \citep{todorov2012mujoco} at 1:4 Froude scale. Gravity-driven motion keeps its dynamics under Froude scaling when lengths scale by 1/4, times by 1/2, speeds by 1/2 and forces by 1/64. Every number in this paper is converted to full scale, except computation times, which are wall-clock; Table~\ref{tab:scaling} lists the scaling and Table~\ref{tab:params} the main parameters. A 9.6-m joint hangs from the elevator on a ball joint and swings with a period of 5.07\,s, the compound-pendulum value. The travelling block moves on a vertical slide driven by a position actuator, and its height is available to the controller, as it would be from a drawworks encoder. The box is held in the rotary at well centre.

\begin{table}[!htbp]
\centering
\caption{Froude scaling between the model and full scale}
\label{tab:scaling}
\small
\begin{tabular}{@{}l l c c@{}}
\toprule
Quantity & Scale factor & Model & Full scale\\
\midrule
Length (joint) & 1/4 & 2.4\,m & 9.6\,m\\
Time (swing period) & 1/2 & 2.54\,s & 5.07\,s\\
Speed & 1/2 & \multicolumn{2}{c}{model = full scale / 2}\\
Mass (heaviest joint) & 1/64 & 21\,kg & 1,343\,kg\\
Force & 1/64 & \multicolumn{2}{c}{model = full scale / 64}\\
Camera rate & 2 & 60\,Hz & 30\,Hz\\
Camera delay & 1/2 & 33\,ms & 66\,ms\\
Simulated-camera noise & 1/4 & 1.5\,mm & 6\,mm\\
Physics time step & 1/2 & 2\,ms & 4\,ms\\
Contact time constant & not scaled & 0.004\,s & --\\
Friction coefficient & not scaled & 0.12--0.5 & --\\
Image size & not scaled & 256$\times$256 & --\\
\bottomrule
\end{tabular}
\tabnote{Froude scaling keeps gravity-driven motion similar when lengths scale by 1/4. The contact model and friction are set in the model and are not scaled, which is why contact speed rather than force leads the comparison (Section~\ref{sec:exp}). Inference time is wall-clock and is not scaled.}
\end{table}

\begin{table}[!htbp]
\centering
\caption{Main parameters of the keypoint network, the filter and the residual policy}
\label{tab:params}
\small
\begin{tabular}{@{}l >{\raggedright\arraybackslash}p{0.62\linewidth}@{}}
\toprule
Item & Setting\\
\midrule
Keypoint network & ResNet-18 (ImageNet), three heatmaps of 64$\times$64\\
Heatmap target & Gaussian, $\sigma$ = 1.5\,px\\
Network training & AdamW, learning rate $10^{-3}$, weight decay $10^{-4}$, one-cycle schedule (10\% warm-up), 30 epochs, batch 64\\
Data & 20,000 training / 2,000 validation / 2,000 test stereo pairs\\
Filter measurement noise & 6\,mm (not retuned for the network)\\
Residual policy & MLP 64$\times$64, PPO (Stable-Baselines3)\\
PPO settings & 8 environments $\times$ 256 steps, batch 512, 10 epochs, learning rate $3{\times}10^{-4}$, $\gamma$ 0.99, $\lambda$ 0.95, clip 0.2\\
PPO training & 600,000 decisions (8,494 episodes), seed 0\\
\bottomrule
\end{tabular}
\tabnote{Controller timings and thresholds are given at full scale in Section~\ref{sec:method}.}
\end{table}

Nine sizes are modelled with their body and tool-joint diameters: 2$\frac{3}{8}$-in. tubing; 3$\frac{1}{2}$-, 5- and 6$\frac{5}{8}$-in. drillpipe; and 7-, 9$\frac{5}{8}$-, 13$\frac{3}{8}$-, 18$\frac{5}{8}$- and 20-in. casing, up to 1,343\,kg per joint. Pin nose and box mouth carry a taper. Threads are not modelled, so a stab is judged by the pin entering the box and the shoulder coming down.

\begin{figure}[!htbp]
\centering
\includegraphics[width=\textwidth]{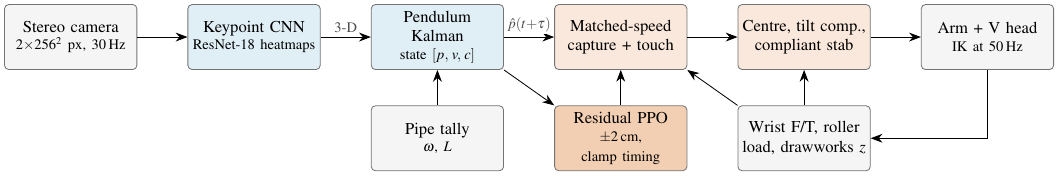}
\figcap{Overview of StabBot}{The vision network gives 3-D landmarks, from which the pendulum filter predicts the pipe. The model-based controller captures, centres and stabs, and the residual policy adds a small, smoothed correction during capture. The rig inputs are ones that drill floor robots already have, namely the pipe tally, the drawworks height, cameras and a wrist force sensor.}
\label{fig:system}
\end{figure}

The robot has the kinematics of a UR10e from MuJoCo Menagerie \citep{zakka2022menagerie}, scaled with the scene, with joint position servos tracking a damped least-squares inverse kinematics solution \citep{wampler1986manipulator} at 50\,Hz and a six-axis force-torque sensor at the wrist. It stands in for an industrial drill floor manipulator. Two fixed cameras on the far side of the well, 90$^\circ$ apart, view the pipe, the head and the stick-up at 256$\times$256 pixels. Fig.~\ref{fig:system} shows how the parts connect.

\subsection{Guide head}
\begin{figure}[!htbp]
\centering
\begin{minipage}[b]{0.255\textwidth}\centering
\includegraphics[width=\linewidth]{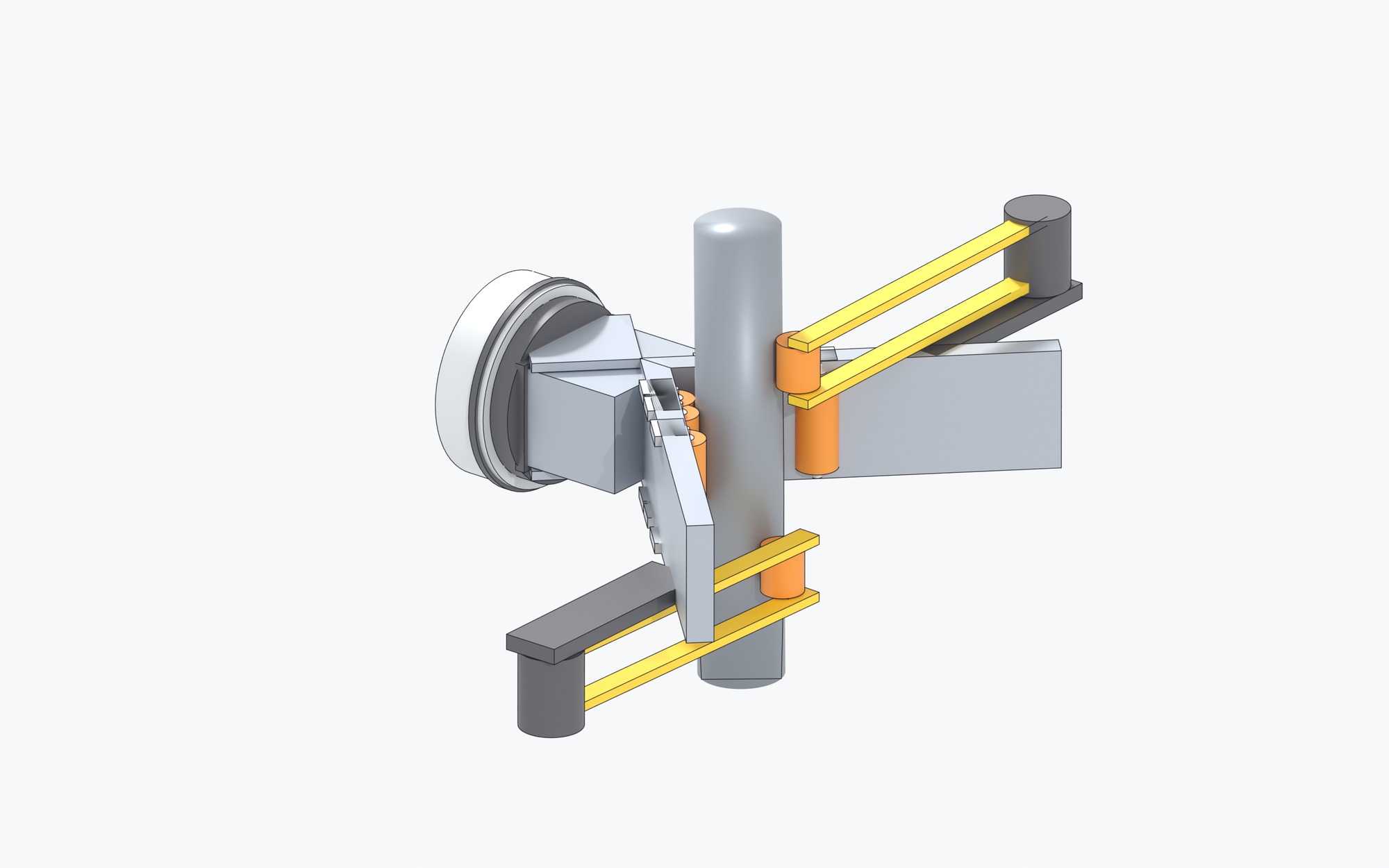}\\[-1pt]{\footnotesize (a) clamped on 5-in. drillpipe}\\[4pt]
\includegraphics[width=\linewidth]{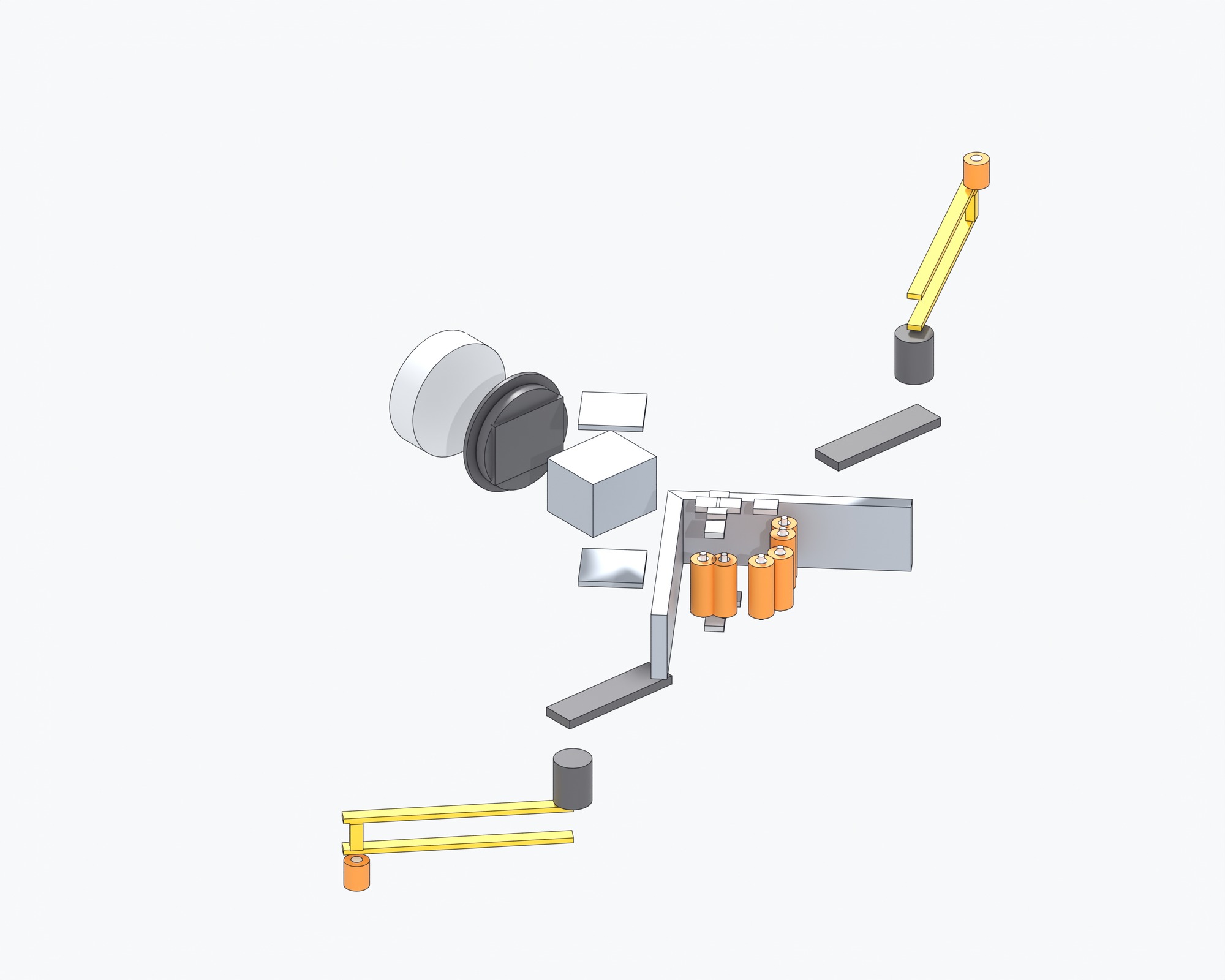}\\[-1pt]{\footnotesize (b) exploded, clamps open}
\end{minipage}\hfill
\begin{minipage}[b]{0.43\textwidth}\centering
\includegraphics[width=\linewidth]{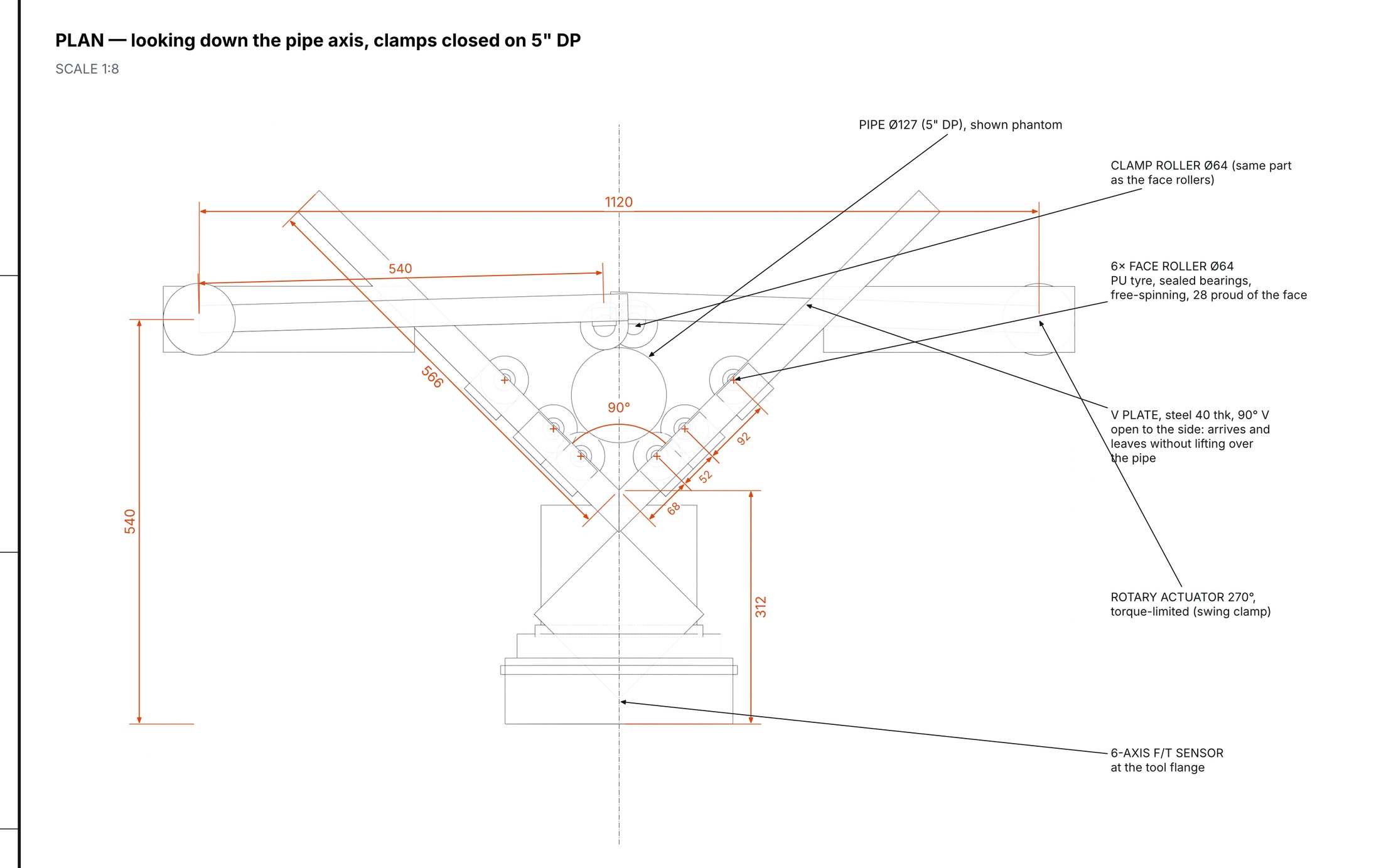}\\[-1pt]{\footnotesize (c) plan view, dimensions in mm}
\end{minipage}\hfill
\begin{minipage}[b]{0.29\textwidth}\centering
\includegraphics[width=\linewidth]{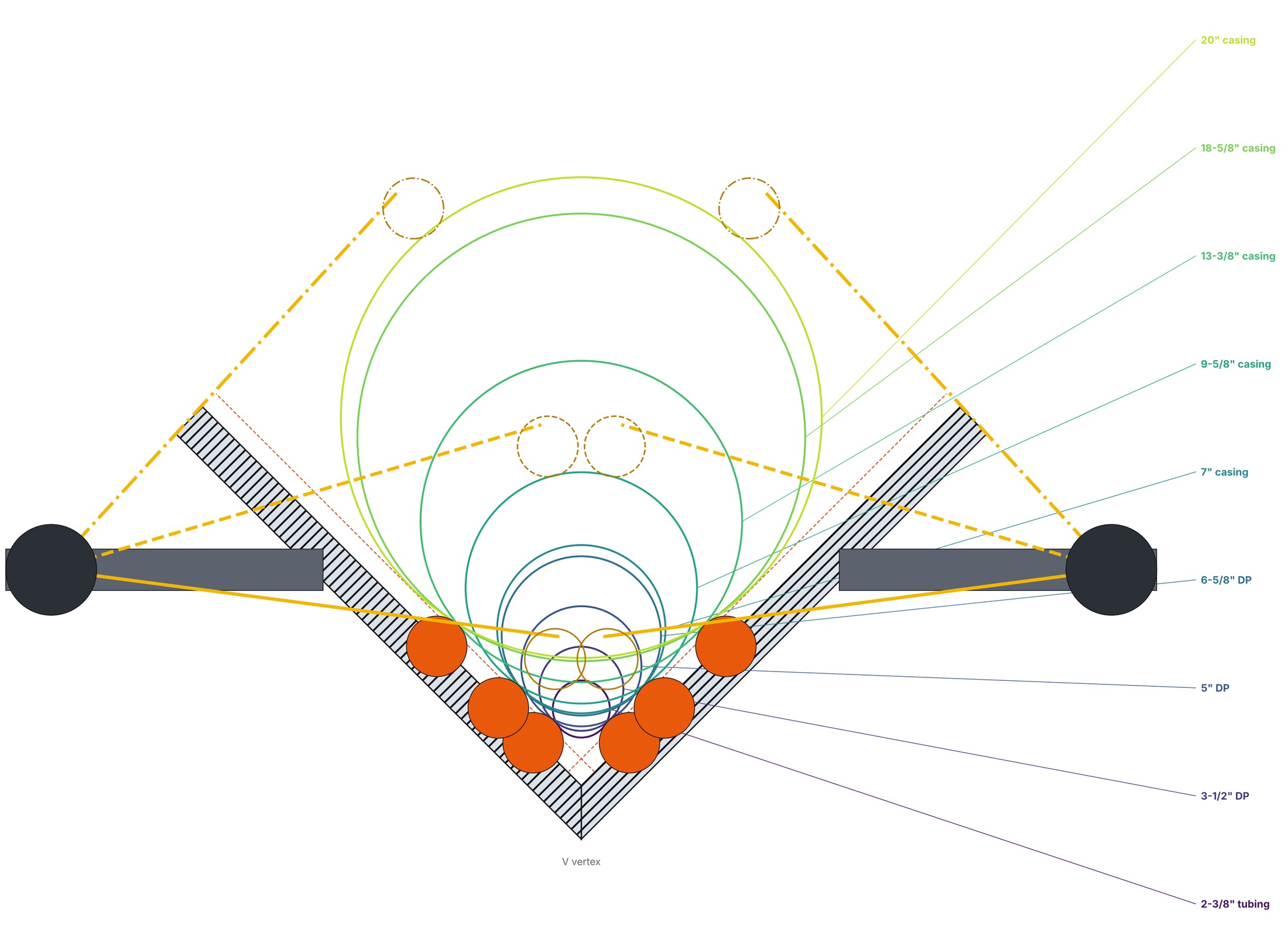}\\[-1pt]{\footnotesize (d) exact section, nine sizes seated}
\end{minipage}
\figcap{The guide head designed for this work}{(a) Rendered design with the clamps closed on 5-in. drillpipe. (b) Exploded view: flange adapter and 6-axis F/T sensor, mount block, V plates, six face rollers on standoffs, outrigger brackets, two 270$^\circ$ torque-limited rotary actuators and clamp arms ending in rollers of the same part. (c) General arrangement. (d) Section through the rollers with every size seated, from 2$\frac{3}{8}$-in. tubing to 20-in. casing: each touches rollers only, and the clamps close on every diameter. The full set of four A3 sheets is in the repository.}
\label{fig:head}
\end{figure}

The head (Fig.~\ref{fig:head}) is an open 90$^\circ$ V with 566-mm faces, 184\,mm high, carrying three free rollers per face (64-mm diameter, 28\,mm proud of the plate) and two swing clamps, 540\,mm long, driven by torque-limited rotary actuators, each ending in a roller of the same part number. The V opens toward the pipe, so the head arrives from the side and leaves after the stab without lifting over the pipe. Roller stations, plate angle and clamp pivots were sized together for all nine diameters under three rules: (i)~the pipe touches only rollers; (ii)~every other part stays at least 20\,mm from the pipe surface, checked with exact geometric distances for every size; (iii)~the clamps push the pipe toward the V vertex at every size, with a worst push angle of 39$^\circ$. Because the rollers turn freely, a held joint can still spin: the torque to turn it was 3.2--8.4\,N$\cdot$m, 16 to 60 times less than with locked rollers. A capture test of every size gave an envelope of 0.8\,cm (2$\frac{3}{8}$-in. tubing and 7-in. casing) to 1.6\,cm (drillpipe) at the box, which sets the accuracy the stab needs. Fig.~\ref{fig:head}d shows the same head on all nine sizes.

\section{Method}\label{sec:method}
\begin{figure}[!htbp]
\centering
\includegraphics[width=\textwidth]{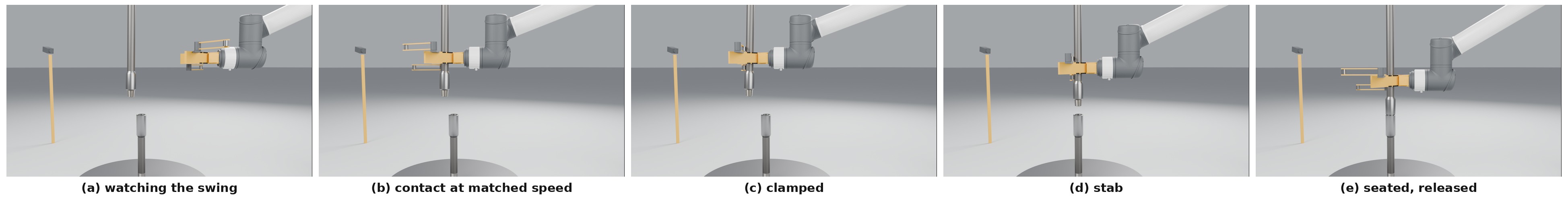}\\[1pt]
\includegraphics[width=\textwidth]{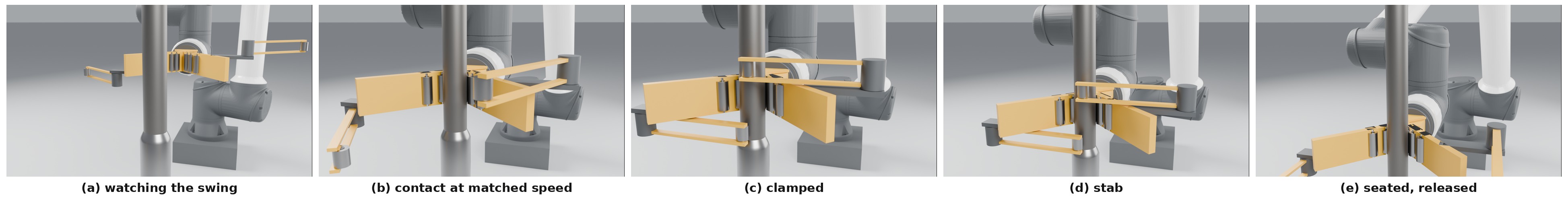}
\figcap{One job with StabBot}{5-in. drillpipe, 20-cm swing at the pin, seen from the side (top) and from beyond the pipe (bottom). (a) The robot watches the swing; (b) the V meets the pipe at its speed; (c) the clamps close; (d) the block slacks off and the pin enters the box; (e) the clamps open and the head backs off. The frames replay the recorded physics in a plain studio setting.}
\label{fig:sequence}
\end{figure}

A job has four phases: \emph{capture} (bring the V onto the swinging pipe and close the clamps), \emph{centre} (bring the held pipe to rest over the box), \emph{stab} (slack off the block while the head follows and lets the pin find the box) and \emph{release}. Fig.~\ref{fig:sequence} shows one job.

\subsection{Swing estimation}
For each horizontal axis a Kalman filter \citep{kalman1960new} tracks $\mathbf{x}=[p,\,v,\,c]^\top$, the position and velocity of the pipe axis at head height and the rest position $c$ about which it swings. The hanging joint obeys $\ddot p=-\omega^2(p-c)$, $\dot c=0$, with $\omega$ computed from the pipe tally (joint length and mass distribution). The exact discretisation over a step $\Delta t$ is
\begin{equation}
\begin{aligned}
p_{k+1} &= c_k + (p_k-c_k)\cos\omega\Delta t + \tfrac{v_k}{\omega}\sin\omega\Delta t,\\
v_{k+1} &= -\omega(p_k-c_k)\sin\omega\Delta t + v_k\cos\omega\Delta t,\qquad c_{k+1}=c_k .
\end{aligned}
\label{eq:kf}
\end{equation}
\begin{where}
$p_k$, $v_k$ & Position and velocity of the pipe axis at head height at step $k$ & m, m/s\\
$c_k$ & Rest position about which the joint swings & m\\
$\omega$ & Angular frequency of the swing, from the pipe tally & rad/s\\
$\Delta t$ & Time between filter steps & s\\
$k$ & Filter step index & --\\
\end{where}
Camera samples update the filter at their capture time, so the filter also absorbs the camera delay, and \eqref{eq:kf} predicts the pipe at any future time in closed form. The rest position is where the free pipe would hang below the elevator, and it is used for the tilt compensation in Section~\ref{sec:tilt}. In the benchmark $\omega$ carries a random error of up to $\pm$3\%.

\subsection{Velocity-matched capture}
The arm waits with the V vertex 1.08\,m from well centre, clear of any swing, and watches for 0.8\,s. Over the next 2.0\,s it blends into shadowing the predicted pipe: the V target is the pose that would seat the pipe predicted $\tau=0.12$\,s ahead (to cover servo lag), backed off by a stand-off $s$ along the approach direction $\mathbf n$,
\begin{equation}
\mathbf y^*(t)=\hat{\mathbf p}(t+\tau)+s(t)\,\mathbf n .
\label{eq:capture}
\end{equation}
\begin{where}
$\mathbf y^*(t)$ & Target position of the V seat at time $t$ & m\\
$\hat{\mathbf p}(t+\tau)$ & Pipe position predicted $\tau$ ahead with \eqref{eq:kf} & m\\
$\tau$ & Look-ahead that covers servo lag (0.12\,s) & s\\
$s(t)$ & Stand-off along the approach direction & m\\
$\mathbf n$ & Unit vector along the approach direction & --\\
\end{where}
Over the following 1.2\,s, $s$ follows a minimum-jerk profile from 12\,cm down to an 8-mm preload, so the V reaches the pipe moving with it. The clamps pre-swing to 60$^\circ$ short of contact and close 0.8\,s after the first face-roller contact, when the V already rides the pipe.

\subsection{Guarded final approach}
If the face rollers have not felt the pipe 0.4\,s after close-in, the V keeps advancing at 2\,cm/s, up to 6\,cm, until the roller load exceeds 64\,N, and holds that depth \citep{lozanoperez1984automatic}. Capture therefore ends when contact is sensed at the rollers, so a small error in the vision estimate cannot leave the V hovering just short of the pipe.

\subsection{Tilt compensation}\label{sec:tilt}
If the elevator hangs off well centre, the free pipe is tilted, and holding it at well centre at head height leaves the pin off the box. The rest estimate $c$, frozen at first contact, gives the elevator offset $o$. With the elevator, head and pin nose at heights $z_e$, $z_h$ and $z_p$, the head holds the pipe at
\begin{equation}
x_h = o\,\frac{r}{1+r},\qquad r=\frac{z_h-z_p}{z_e-z_h},
\label{eq:tilt}
\end{equation}
\begin{where}
$x_h$ & Hold position of the pipe at head height, measured from well centre & m\\
$o$ & Offset of the elevator from well centre & m\\
$r$ & Ratio of the head-to-pin and elevator-to-head heights & --\\
$z_e$, $z_h$, $z_p$ & Heights of the elevator, the head and the pin nose & m\\
\end{where}
which puts the pin nose over the box. It is applied when $|o|$ exceeds 3\,cm, about three times the estimation error.

\subsection{Centring, stab and release}
After capture the head brings the pipe to its hold point along a quintic trajectory of 3.2\,s that starts at the current head velocity. The pipe is settled when the mean of the last ten camera samples lies within 6\,mm of the hold point. The block then slacks off at 0.1\,m/s and the head follows it down from the drawworks height. Within 2\,cm of the box mouth an admittance law \citep{hogan1985impedance} on the wrist force, with a 96-N deadband, lets the pipe push the head sideways, as a floorhand's hand gives, so that the taper finishes the alignment. With the pin half way in, the clamps open, the head backs off, and the block continues until the shoulder seats. A block lagging more than 2.4\,cm behind its command means that the pin is resting on the box face, and the job is then stopped and counted as failed.

\subsection{Stereo keypoint network}
A generator renders 20,000 training, 2,000 validation and 2,000 test stereo pairs with exact labels: random size, tilt up to 2$^\circ$, slack-off up to 40\,cm, robot pose (ready, seated, or shadowing with open clamps, which gives realistic occlusion), colours, lighting and, in training only, camera-mount jitter of 4\,cm and 1$^\circ$ \citep{tobin2017domain}. Three keypoints are labelled per image: the pipe axis at head height, the pin nose and the box mouth. An ImageNet-pretrained ResNet-18 \citep{he2016resnet} feeds three transposed convolutions and one 64$\times$64 heatmap per keypoint \citep{xiao2018simple}, read with a soft-argmax \citep{sun2018integral}. The loss is the L1 keypoint error plus 0.1 times the cross-entropy to a small Gaussian at the label. The final network (v2) also trains with random occluding rectangles, glare spots and motion blur that differ from the test patterns. The pipe point is solved on its known plane at head height; pin and box are triangulated by direct linear transformation \citep{hartley2004multiple}.

\subsection{Residual policy}
Every 0.08\,s during capture the policy adds a correction to the controller: a shift of the V target of up to $\pm$2\,cm along and across the approach, passed through a critically damped filter with a 0.1-s time constant, and a clamp-timing signal \citep{johannink2019residual,silver2018residual}. Its 15 observations are the robot's own: predicted pipe position and velocity relative to the seat, commanded head velocity, roller load, clamp command and angle, a seat-contact flag, elapsed time, pipe radius, stand-off and the filtered correction. The reward for decision $k$ is
\begin{equation}
r_k = -0.05\,F_k - 0.01 - 0.10\,u_c\,\delta_k^{c} + 10\,\delta_k^{s} - 10\,\delta_k^{f} - 0.5\,F_{\max}\,\delta_k^{e},
\label{eq:reward}
\end{equation}
\begin{where}
$r_k$ & Reward for decision $k$ & --\\
$F_k$ & Peak head--pipe force during decision step $k$ & kN\\
$u_c$ & Contact speed at the first roller contact & cm/s\\
$F_{\max}$ & Peak head--pipe force over the episode & kN\\
$\delta_k^{c}$, $\delta_k^{s}$ & 1 at first contact / when the pipe is centred, otherwise 0 & --\\
$\delta_k^{f}$, $\delta_k^{e}$ & 1 on a miss or non-roller contact / at the last step, otherwise 0 & --\\
\end{where}
so that time, first impact, every force peak and the outcome are all priced. A 64$\times$64 policy is trained with PPO \citep{schulman2017ppo,raffin2021sb3} for 600,000 decisions on jobs from different seeds than the benchmark.

\subsection{Baselines}
\emph{Static guide} stands for a conventional stabbing guide: the V is parked over well centre and the clamps close when the swinging pipe strikes the rollers. \emph{Reactive} follows the latest camera sample with no prediction. \emph{Reactive+guarded} adds the guarded approach, to separate prediction from touch. \emph{Swing-aware} is StabBot without the residual policy. All use the same head, centring, stab and release. The static guide represents current practice, in which the guide waits at well centre for the swinging joint; a floorhand steadying the pipe by hand is not modelled. Because it does not use the camera to capture, its centring check uses the simulated camera, and its centring forces are small (median 0.02\,kN), so this choice has little effect on the comparison.

\section{Experimental Design}\label{sec:exp}
\emph{Benchmark.} Each of 200 jobs draws a size (uniform over nine), a swing of 5--25\,cm at the pin in any direction, an elliptical component, a starting phase, a $\pm$3\% error in $\omega$ and a camera-noise seed. Every method runs every job, so all comparisons are paired. A job succeeds when the pin is seated (shoulder gap under 8\,mm, lateral offset under 12\,mm) and the pipe touched nothing but rollers on the head and nothing on the robot. In the main benchmark the keypoint network is in the loop at 30\,Hz with a 66-ms delay.

\emph{Metrics.} Contact speed is the relative horizontal speed of pipe and head at first roller contact. It leads the comparison because it depends on the controller and the swing, not on the contact stiffness chosen in the simulator, and the kinetic energy of the relative motion at first contact grows with its square. Peak force is the largest net horizontal head--pipe force during capture and centring. Peak wrist force is the largest change in the wrist reading. Time to seat runs to the shoulder seating.

\emph{Statistics.} Success rates carry Wilson 95\% intervals \citep{wilson1927probable}. Paired differences are tested with the two-sided Wilcoxon signed-rank test \citep{wilcoxon1945individual} (normal approximation with tie correction, SciPy 1.18 \citep{virtanen2020scipy}) on the per-job differences; jobs in which a method never touched the pipe have no contact speed and are left out of that test. Effect sizes are ratios of medians with bootstrap 95\% intervals (5,000 resamples), and the $p$ values of the ablation are checked with a Holm correction.

\emph{Robustness.} On the first 100 jobs, one error at a time is added at three levels with the controllers unchanged: noise of 12/20/30\,mm added to the network's estimate, delay 133/200/300\,ms (nominal 66\,ms), 30/60/85\% of frames lost, calibration offset 5/10/20\,mm, frequency error 5/10/15\%, elevator 5/10/20\,cm off centre, and swings of 25--30, 30--35 and 35--40\,cm. The keypoint network stays in the loop throughout, and the added noise is Gaussian. The fixed guide does not track the pipe, so its curves come from the same jobs with the simulated camera.

\emph{Unseen conditions.} 450 new scenes are rendered nominally and under 14 conditions the network never saw, each scene under every condition.

\section{Results}\label{sec:results}

\subsection{Main benchmark}
\begin{figure}[!htbp]
\centering
\includegraphics[width=\linewidth]{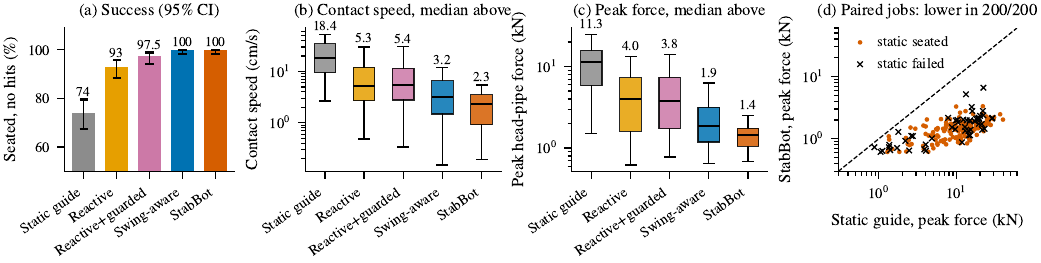}
\figcap{Benchmark over 200 paired jobs}{(a) Jobs seated with the pipe touching only rollers, Wilson 95\% intervals. (b, c) Contact speed and peak head--pipe force (box: interquartile range; whiskers: 5th--95th percentile; log scale; median printed). (d) Peak force per job, fixed guide against StabBot: every point lies below the diagonal, including the 52 jobs the fixed guide failed ($\times$).}
\label{fig:bench}
\end{figure}

Table~\ref{tab:main} and Fig.~\ref{fig:bench} give the result. StabBot seated all 200 joints at every size and every swing amplitude, with nothing but rollers touching the pipe. The fixed guide seated 74.0\%: it never captured the pipe in 21 jobs and in 31 more it seated the pin only after the pipe struck steel on the head. Its success dropped to 60\% for swings of 20--25\,cm, and to 57\% on 20-in. casing.

\begin{figure}[!htbp]
\centering
\includegraphics[width=\linewidth]{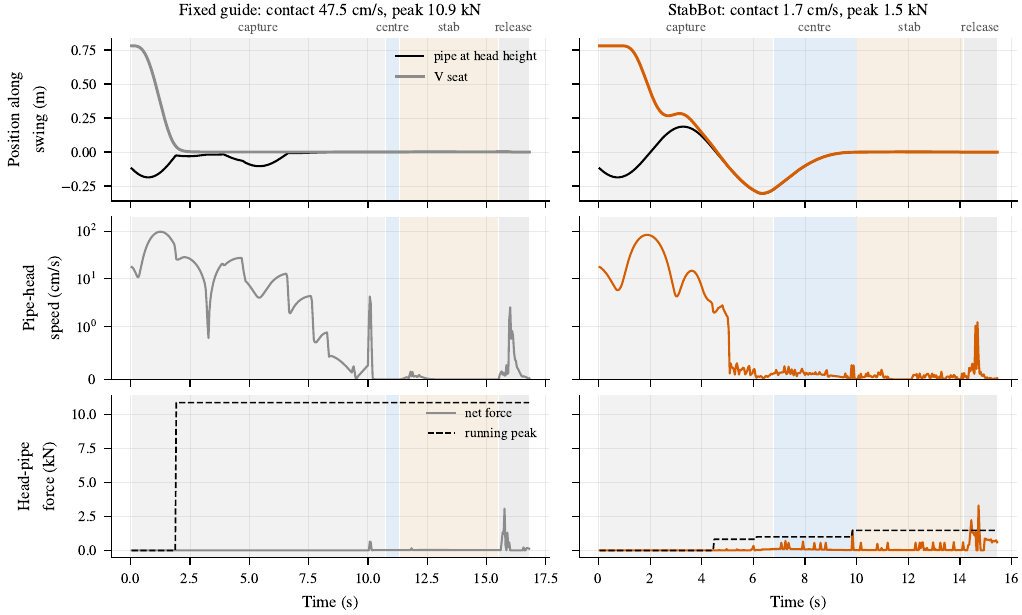}
\figcap{One recorded job, fixed guide against StabBot}{5-in. drillpipe, 20-cm swing at the pin, keypoint network in the loop. Top: position along the swing direction of the pipe axis at head height and of the V seat. Middle: pipe--seat relative speed (symmetric log scale). Bottom: net head--pipe force sampled at 25 frames/s and the running peak over capture and centring (dashed), which also records impacts shorter than a frame. Shading marks the job phases. The fixed guide waits at well centre and is struck at 47.5\,cm/s; StabBot follows the pipe, reaches it at 1.7\,cm/s and holds it while it centres.}
\label{fig:timeseries}
\end{figure}

Compared with the fixed guide, StabBot lowered median contact speed from 18.4 to 2.3\,cm/s and median peak force from 11.3 to 1.4\,kN, a factor of 7.8 for both (bootstrap 95\% intervals 6.4--10.0 and 7.1--8.7), and the 95th-percentile force from 24.7 to 2.5\,kN. Paired job by job, its peak force was lower in all 200 jobs (Fig.~\ref{fig:bench}d; Wilcoxon $p<10^{-30}$) and its contact speed in 173 of the 182 jobs in which the fixed guide touched the pipe ($p<10^{-30}$). The relative kinetic energy at first contact, which does not depend on the contact model, was a median 95 times lower per job. Peak wrist force fell by 66\% ($p<10^{-28}$). On the 148 jobs that both methods seated cleanly, the fixed guide was 1.8\,s faster (median of the per-job differences) because it does not observe the swing first, but it failed one job in four overall. Table~\ref{tab:breakdown} breaks the result down by size and swing amplitude, and Fig.~\ref{fig:timeseries} follows one job with both methods.

\begin{table}[!htbp]
\centering
\caption{Main benchmark: 200 paired jobs, keypoint network in the loop}
\label{tab:main}
\small
\begin{tabular}{@{}l c c c c@{}}
\toprule
Method & Seated, no hits & Contact speed, cm/s & Peak force, kN & Time to\\
 & (95\% CI) & (med / p95) & (med / p95) & seat, s\\
\midrule
Static guide$^*$ & 74.0\% (67.5--79.6) & 18.4 / 53.4 & 11.3 / 24.7 & 12.3\\
Reactive & 93.0\% (88.6--95.8) & 5.3 / 30.5 & 4.0 / 13.4 & 13.6\\
Reactive+guarded & 97.5\% (94.3--98.9) & 5.4 / 30.1 & 3.8 / 14.1 & 13.6\\
Swing-aware & 100\% (98.1--100) & 3.2 / 11.8 & 1.9 / 6.3 & 14.6\\
\textbf{StabBot} & \textbf{100\% (98.1--100)} & \textbf{2.3 / 5.5} & \textbf{1.4 / 2.5} & 14.1\\
\midrule
StabBot vs static guide (paired) & $+26$ pts & $-87\%$ & $-87\%$ & $+1.8$\\
\bottomrule
\end{tabular}
\tabnote{Every method ran the same 200 random jobs. CI: Wilson 95\% interval; med: median; p95: 95th percentile. $^*$The static guide does not use the camera to capture; its row is the simulated-camera run of the same jobs. Time to seat is the median over jobs seated with no hits. Contact speed of the static guide is over the 182 jobs in which the pipe touched the head.}
\end{table}

\begin{table}[!htbp]
\centering
\caption{Benchmark by pipe size and by swing amplitude, fixed guide against StabBot}
\label{tab:breakdown}
\small
\begin{tabular}{@{}l c c c c@{}}
\toprule
Group & Jobs & Seated, no hits & Contact speed, cm/s & Peak force, kN\\
 & & (fixed / StabBot) & (median, fixed / StabBot) & (median, fixed / StabBot)\\
\midrule
2$\frac{3}{8}$-in. tubing & 25 & 15 / 25 & 20.0 / 1.4 & 1.8 / 0.69\\
3$\frac{1}{2}$-in. drillpipe & 25 & 19 / 25 & 17.4 / 0.8 & 6.9 / 1.34\\
5-in. drillpipe & 23 & 19 / 23 & 21.7 / 1.8 & 8.5 / 0.97\\
6$\frac{5}{8}$-in. drillpipe & 23 & 21 / 23 & 24.0 / 2.8 & 12.0 / 1.33\\
7-in. casing & 23 & 17 / 23 & 17.6 / 3.0 & 10.3 / 1.41\\
9$\frac{5}{8}$-in. casing & 17 & 15 / 17 & 18.1 / 1.8 & 12.8 / 1.60\\
13$\frac{3}{8}$-in. casing & 19 & 14 / 19 & 28.5 / 2.6 & 14.0 / 1.74\\
18$\frac{5}{8}$-in. casing & 22 & 15 / 22 & 16.2 / 2.4 & 19.3 / 1.74\\
20-in. casing & 23 & 13 / 23 & 19.0 / 2.3 & 17.8 / 2.03\\
\midrule
Swing 5--10\,cm & 43 & 33 / 43 & 13.4 / 1.8 & 7.1 / 1.11\\
Swing 10--15\,cm & 51 & 40 / 51 & 13.2 / 2.1 & 12.5 / 1.43\\
Swing 15--20\,cm & 58 & 46 / 58 & 20.4 / 2.7 & 13.0 / 1.58\\
Swing 20--25\,cm & 48 & 29 / 48 & 23.1 / 3.3 & 12.5 / 1.53\\
\bottomrule
\end{tabular}
\tabnote{Same 200 jobs as Table~\ref{tab:main}; swing is the amplitude at the pin. StabBot seated every job in every group, and its median peak force stayed between 0.7 and 2.0\,kN across sizes, while that of the fixed guide grew with pipe weight.}
\end{table}

\subsection{Contribution of each component}
Table~\ref{tab:ablation} adds the components one at a time. The guarded approach raised reactive tracking from 93.0 to 97.5\% success (12 jobs gained, 3 lost; McNemar test \citep{mcnemar1947note}, $p\approx0.035$), with a small change in force ($-4\%$) and contact speed ($+7\%$). Swing-aware capture differs from reactive+guarded in prediction, velocity matching and the clamp timing that goes with them (a 0.8-s watch, pre-swung clamps that close once the V rides the pipe); together these cut contact speed by 41\% and peak force by 51\% and removed the remaining failures. The residual policy took a further 27\% off contact speed and 22\% off peak force and saved 0.5\,s per connection. In 56\% of swing-aware jobs the peak force occurred before the clamps closed, compared with 24\% with the policy, whose remaining peaks fell at clamp landing or during centring, both near 1\,kN.

\begin{table}[!htbp]
\centering
\caption{Paired effect of each component (200 jobs)}
\label{tab:ablation}
\small
\begin{tabular}{@{}l c c c@{}}
\toprule
Step & Contact speed & Peak force & Success\\
\midrule
+ guarded approach & $+7\%$ ($p$ = 0.02) & $-4\%$ ($p$ = 0.02) & 93.0 $\to$ 97.5\%\\
+ prediction and matched speed & $\mathbf{-41\%}$ ($p$ = $1{\times}10^{-7}$) & $\mathbf{-51\%}$ ($p$ = $3{\times}10^{-16}$) & 97.5 $\to$ 100\%\\
+ residual RL & $\mathbf{-27\%}$ ($p$ = $3{\times}10^{-8}$) & $\mathbf{-22\%}$ ($p$ = $4{\times}10^{-19}$) & 100 $\to$ 100\%\\
\midrule
Static guide $\to$ StabBot & $\mathbf{-87\%}$ ($p$ = $9{\times}10^{-31}$) & $\mathbf{-87\%}$ ($p$ = $1{\times}10^{-34}$) & 74 $\to$ 100\%\\
\bottomrule
\end{tabular}
\tabnote{Each row adds one component to the controller of the row above and is compared with it on the same jobs. Percentages are changes in the median over the jobs in which both controllers touched the pipe. $p$ values are from the two-sided Wilcoxon signed-rank test on the per-job differences, and all eight stay below 0.05 after Holm correction.}
\end{table}

\subsection{Robustness}
\begin{figure}[!htbp]
\centering
\includegraphics[width=\linewidth]{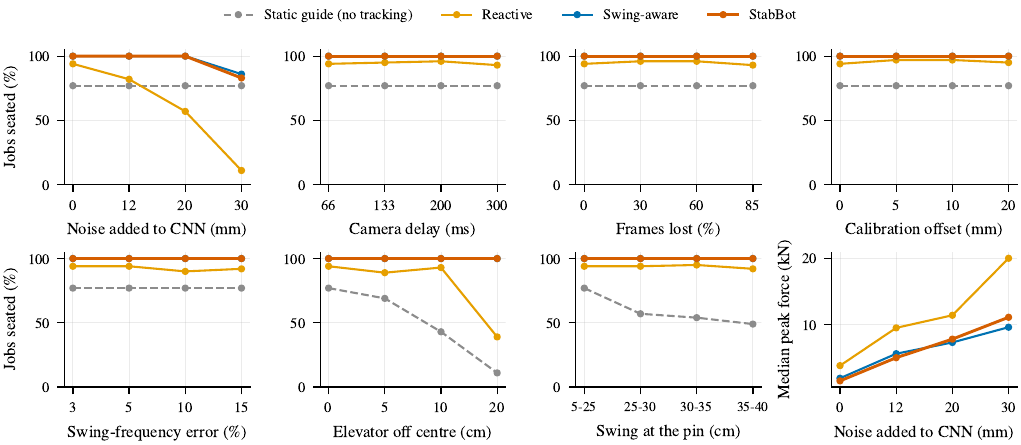}
\figcap{Robustness of the full system with the keypoint network in the loop}{First 100 jobs, one error at a time with the controllers unchanged. The two swing-aware methods, whose curves overlap in most panels, seat 100\% of joints through every error except 30\,mm of added noise; the last panel shows that noise raises their force while reactive tracking misses. The fixed guide (dashed) does not track the pipe. Swing levels are ranges at the pin.}
\label{fig:robust}
\end{figure}

Fig.~\ref{fig:robust} adds one error at a time to the full system, keypoint network in the loop. Both swing-aware methods seated 100\% of jobs at every level of camera delay (up to 300\,ms), frame loss (up to 85\%), calibration offset (20\,mm), frequency error (15\%), elevator offset (20\,cm) and swing (35--40\,cm), and with up to 20\,mm of noise added to the network's estimate; with 30\,mm added they seated 83--86\%. StabBot's median peak force stayed between 1.4 and 2.0\,kN under delay, calibration, frequency and elevator errors, against 1.5\,kN nominal on these jobs, and reached 2.7\,kN for 35--40-cm swings, 4.7\,kN with 85\% of frames lost and 7.8\,kN with 20\,mm of added noise. Reactive tracking, which follows each sample, fell to 57\% with 20\,mm of added noise (11\% at 30\,mm), with 11.4-kN median forces, and to 39\% at a 20-cm elevator offset. We attribute the difference to the filter, which averages samples against a model of how a hanging pipe moves, predicts across lost frames and delay, and estimates where the elevator is. The fixed guide does not track the pipe and is unaffected by camera errors, but fell to 11\% at a 20-cm elevator offset and 49\% at 35--40-cm swings. Repeating the test with the simulated camera gave the same pattern (99--100\% for the swing-aware methods except at 30\,mm of noise). Under the largest errors StabBot and swing-aware performed alike, with StabBot keeping the force lower under a 20-mm calibration offset (2.0 against 4.3\,kN).

\subsection{Off-centre elevator}
\begin{figure}[!htbp]
\centering
\includegraphics[width=0.8\linewidth]{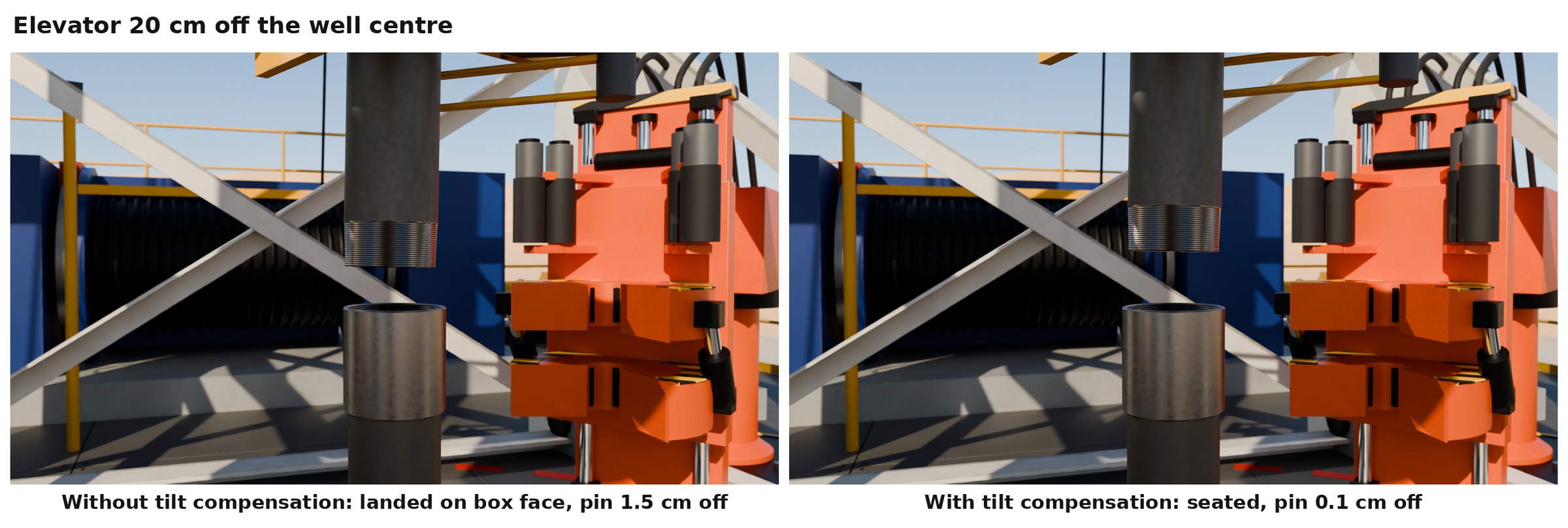}
\figcap{Elevator 20\,cm off centre, same job}{Without tilt compensation the pin lands on the box face 1.5\,cm off (left); with it the joint seats 0.1\,cm off (right).}
\label{fig:tilt}
\end{figure}

With the elevator 20\,cm off centre, the pipe hung tilted and the pin arrived 1.4--1.5\,cm off the box, outside the capture envelope of six of the nine sizes. Without compensation, swing-aware capture seated 43\% of these jobs. With \eqref{eq:tilt}, both swing-aware methods seated 100\%, with the pin 0.1--0.3\,cm off the box (Fig.~\ref{fig:tilt}). The rest estimate was within about 1\,cm of the true elevator position and never triggered a correction on the nominal jobs.

\subsection{Vision}
On 2,000 held-out scenes the network's 3-D error was 0.4\,mm median (1.1\,mm p95) for the pipe at head height, 0.7/1.6\,mm for the pin nose and 0.5/1.5\,mm for the box mouth (Table~\ref{tab:vision}). The pin-to-box offset, which decides the stab, was within 6\,mm in 99.9\% of scenes, against a smallest capture envelope of 8\,mm. One stereo pair takes 2.3\,ms on a laptop RTX 4060, 14 times faster than the 30\,Hz the controller needs. In the loop the network's median error was 0.7\,mm, about a tenth of that of the simulated camera used as a design reference.

\begin{table}[!htbp]
\centering
\caption{Keypoint network on 2,000 held-out scenes (3-D error, mm)}
\label{tab:vision}
\small
\begin{tabular}{@{}l c c c@{}}
\toprule
Landmark & Median & p95 & Within 6\,mm\\
\midrule
Pipe axis at head height & 0.4 & 1.1 & --\\
Pin nose & 0.7 & 1.6 & --\\
Box mouth & 0.5 & 1.5 & --\\
Pin-to-box offset & 0.7 & 1.8 & 99.9\%\\
\bottomrule
\end{tabular}
\tabnote{Error is the distance between the predicted and the true 3-D landmark; p95: 95th percentile. The pin-to-box offset is the one that decides the stab, and its 6-mm threshold lies inside the smallest capture envelope (8\,mm). Inference: 2.3\,ms per stereo pair (426 pairs/s) on a laptop RTX 4060 GPU.}
\end{table}

\begin{figure}[!htbp]
\centering
\begin{minipage}[b]{0.6\textwidth}
\includegraphics[width=\linewidth]{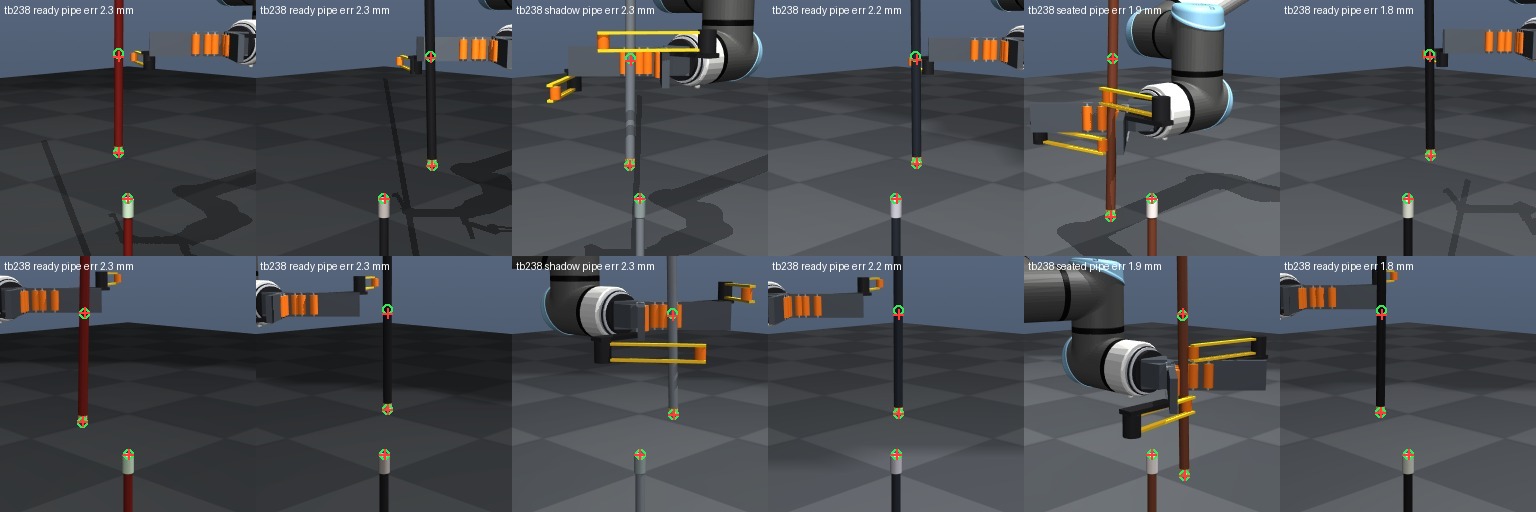}\\[2pt]
\includegraphics[width=\linewidth]{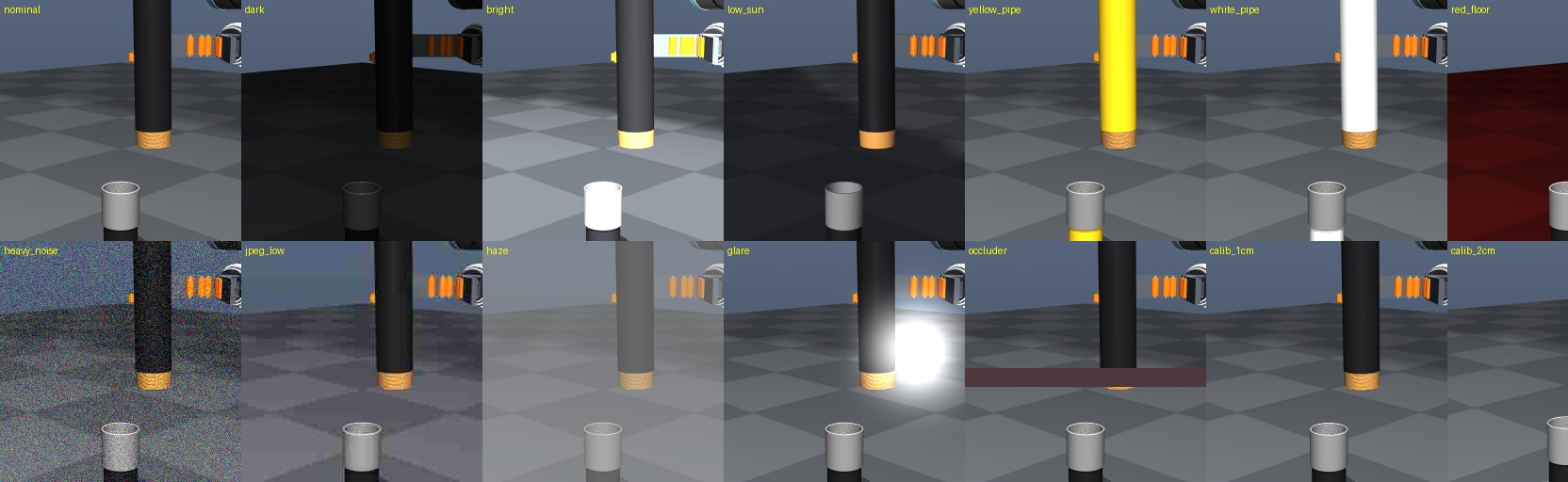}
\end{minipage}\hfill
\begin{minipage}[b]{0.38\textwidth}
\includegraphics[width=\linewidth]{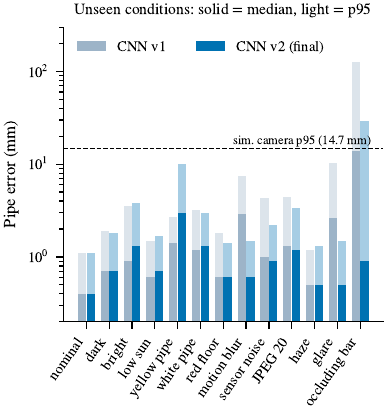}
\end{minipage}
\figcap{Keypoint vision on held-out and unseen scenes}{Top left: predicted (green) and true (red) keypoints on held-out stereo pairs across sizes and robot poses. Bottom left: the same scene under unseen conditions (dark, bright, low sun, yellow and white pipe, red floor, motion blur, sensor noise, JPEG, haze, glare, occluding bar). Right: pipe error for v1 and the final v2 against the 95th-percentile error of the simulated camera used to design the controllers.}
\label{fig:ood}
\end{figure}

Fig.~\ref{fig:ood} shows 12 of the 14 unseen conditions. The other two move the cameras without updating the calibration. The first network (v1) handled changes in light, colour, blur, noise, compression and haze, but glare and an occluding bar produced outliers of tens of centimetres. The final network (v2) was trained with random shapes, glare and blur that were deliberately different from the test patterns, which brought the glare error to 0.5\,mm and the occluded median to 0.9\,mm. When the bar hides a keypoint, the single-image error still has a tail (p95 29\,mm, 88\% of scenes within the simulated camera's p95), and in the loop the filter absorbs it. A calibration error moves the absolute estimate but barely moves the pin-to-box offset (1.5 and 2.6\,mm median), because pin and box shift together. In the loop on 50 jobs, v2 seated 100\% of joints with a bar across the image (peak force 2.4--3.0\,kN, against 8.4--8.5\,kN with v1) and under glare (1.4--1.6\,kN, as nominal), and 98\% with the cameras moved 2\,cm. The v1 network had already seated 100\% in darkness and with motion blur.

The v2 network lost accuracy on a yellow pipe (median 1.4 to 3.0\,mm, p95 2.7 to 10.1\,mm). Retraining on the same scenes with pipes of any paint colour (v3) brought the yellow-pipe error to 0.7/1.5\,mm and the bright-light error, a condition still absent from training, from 1.3/3.8 to 0.9/2.1\,mm, while the error on the original test set was practically unchanged (0.5/1.1\,mm against 0.4/1.1\,mm for v2). In the loop, v3 performed as well as v2. On the 200-job benchmark StabBot again seated 100\% (contact speed 2.4\,cm/s, peak force 1.5/2.5\,kN median/p95), and on 50 jobs it seated 100\% under each of six hard conditions (occluding bar, glare, yellow pipe, bright light, motion blur and a 2-cm camera shift) and at nominal conditions, with median peak forces of 1.3--2.6\,kN. The benchmark results are reported with v2, and v3 is offered as a drop-in replacement for painted pipe.

\subsection{Design lessons}
Two design choices failed, and each led to a change in the final design. \emph{Reward design.} The first residual policy applied its correction as a step and ended the episode at capture. It lowered contact speed by 18\%, but learned to close the clamps early and raised clamp-landing force by 22\%, a case of specification gaming \citep{krakovna2020specification}. Smoothing the action, extending the episode through centring and charging the episode's peak force fixed it. A third policy trained under random sensing errors was worse at nominal conditions (with the network in the loop, contact speed $+14\%$ and peak force $+20\%$) and, with the simulated camera, only marginally better under errors (90 against 86\% at 30-mm noise), so the second policy was kept. Its gain over swing-aware was smaller (contact speed $-16\%$, peak force $-7\%$) than that of the policy reported here. \emph{Final approach.} Before the guarded approach was added, swing-aware capture never missed 2$\frac{3}{8}$-in. tubing with the noisy simulated camera but missed 8 of 25 jobs with the network, although the network was ten times more accurate. With the simulated camera, measurement jitter had pushed the V into contact by chance. With the network, the V hovered millimetres short of the pipe. Closing the last millimetres by touch raised the outcome to 25 of 25.

\section{Core Analysis}\label{sec:core}

\subsection{Why the stab matters}
Every connection made while tripping or running casing begins with a joint swinging from the elevator, and the stab is repeated at every one of them. It is also the step at which people and equipment meet a moving load: a floorhand steadies the pipe by hand inside the red zone, and the pin, the box and the guide take whatever impact the swing delivers \citep{bsee2018energyxxi}. Drill floor robots have taken over the handling of the pipe \citep{raunholt2021field,nabors2025rzr}, but the stab itself is still done by waiting for the pipe at well centre. StabBot changes the stab from an impact into a tracked, matched-speed capture, which removes the hand on the pipe and the shock on the hardware at the same time.

\subsection{Safety and equipment loading}
Table~\ref{tab:core} sets the measured effect per connection against the fixed guide. One fixed-guide job in four ended badly: the guide missed the pipe in 10.5\% of jobs, and in a further 15.5\% the pipe body struck steel on the head before the pin seated. StabBot had neither in any job. Its peak head--pipe force was lower by a factor of 7.8 at the median and 9.9 at the 95th percentile, and lower in every job. The relative kinetic energy at first contact, which does not depend on the contact model, was a median 95 times lower per job, so the pin, the box and the guide receive a light touch instead of a blow. The wrist load fell from 8.1 to 2.8\,kN, which allows a lighter arm and mounting. Because the robot holds the joint from capture to stab-in, no one has to steady the pipe by hand, which takes the floorhand out of the most exposed task at the well centre.

\begin{table}[!htbp]
\centering
\caption{Measured effect per connection, StabBot against the fixed guide}
\label{tab:core}
\small
\begin{tabular}{@{}l c c l@{}}
\toprule
Quantity & Fixed guide & StabBot & Effect\\
\midrule
Capture missed & 10.5\% & 0\% & no re-stab\\
Pipe struck steel, then seated & 15.5\% & 0\% & pipe on rollers only\\
Peak head--pipe force, median / p95 & 11.3 / 24.7\,kN & 1.4 / 2.5\,kN & 7.8$\times$ / 9.9$\times$ lower\\
Contact speed, median & 18.4\,cm/s & 2.3\,cm/s & 7.8$\times$ lower\\
Impact energy at first contact & -- & -- & 95$\times$ lower (median per job)\\
Wrist force, median & 8.1\,kN & 2.8\,kN & 66\% lower\\
Hand on the pipe during the stab & yes & no & floorhand out of the red zone\\
\bottomrule
\end{tabular}
\tabnote{All values from the 200-job paired benchmark with the keypoint network in the loop. Impact energy is the kinetic energy of the relative pipe--head motion at first contact; its ratio is the square of the per-job contact-speed ratio.}
\end{table}

\subsection{Rig time}
A missed capture costs a re-stab. If one re-stab takes $T_r$ seconds, the expected time per connection with each method and the re-stab time above which StabBot is faster are
\begin{equation}
\bar T_G = T_G + P_m\,T_r,\qquad \bar T_S = T_S,\qquad T_r^{*}=\frac{T_S-T_G}{P_m},
\label{eq:time}
\end{equation}
\begin{where}
$\bar T_G$, $\bar T_S$ & Expected time per connection with the fixed guide and with StabBot & s\\
$T_G$, $T_S$ & Median time to seat on the 148 jobs both seated cleanly: fixed guide 12.3\,s, StabBot 14.2\,s & s\\
$P_m$ & Share of jobs in which the fixed guide missed the capture (0.105) & --\\
$T_r$, $T_r^{*}$ & Time of one re-stab, and the value above which StabBot is faster & s\\
\end{where}
which gives $T_r^{*}\approx18$\,s. A re-stab means lifting the joint, letting it settle and trying again, which is likely to take longer than that, so StabBot would save time on average. At $T_r=60$\,s it saves 4.4\,s per connection, or 1.2 rig hours per 1,000 connections, while also removing the re-stabs themselves. The time it spends watching the swing on clean jobs (a median 1.8\,s) is the price of a stab that never misses.

\subsection{Novelty and contributions}
Table~\ref{tab:novelty} summarises what is new in this work and the evidence for each part. To our knowledge, StabBot is the first system that estimates the free swing of a hanging joint online and intercepts it at matched velocity for stabbing; prior drill floor systems wait for the pipe at well centre or guide a stand they already hold (Table~\ref{tab:related}).

\begin{table}[!htbp]
\centering
\caption{Contributions of this work and the evidence for each}
\label{tab:novelty}
\small
\begin{tabular}{@{}p{0.2\linewidth}p{0.42\linewidth}p{0.32\linewidth}@{}}
\toprule
Contribution & What is new & Evidence\\
\midrule
Swing-aware capture & Online pendulum estimate from the pipe tally and the camera; capture at the pipe's own speed, finished by touch & 100\% seated vs 74\%; contact speed and force 7.8$\times$ lower\\
Tilt compensation & Pin aimed over the box from the estimated elevator offset & 43\% $\to$ 100\% seated at a 20-cm offset\\
One head for nine sizes & Open V on free rollers with 20-mm clearance on every other part & 2$\frac{3}{8}$-in. tubing to 20-in. casing, all seated\\
3-D keypoint vision & Pipe, pin and box located from stereo images & 0.4-mm median error, 2.3\,ms per pair\\
Residual learning & Learned correction on top of the model-based capture & a further 22\% lower peak force\\
Paired benchmark & Every method on the same 200 jobs, with robustness and ablation & 100\% under delay, frame loss, calibration and elevator errors\\
\bottomrule
\end{tabular}
\tabnote{Numbers are from Sections~\ref{sec:results} and \ref{sec:core}.}
\end{table}

\subsection{Where StabBot fits}
\emph{Robotised drill floors.} StabBot needs a drill floor arm, the pipe tally, the drawworks height, two fixed cameras with a GPU, and a wrist force sensor. Drill floor robots already provide the arm \citep{raunholt2021field,nabors2025rzr}, and the pipe tally and block height are routine rig data; the cameras, the sensor, the head and the software would be added.

\emph{Casing and completion runs.} Casing is run joint by joint with large, heavy and thread-sensitive connections, where a gentle stab matters most. The fixed guide seated only 57\% of 20-in. casing joints and 60\% of 2$\frac{3}{8}$-in. tubing, and its peak force grew with pipe weight (Table~\ref{tab:breakdown}); StabBot seated every joint of both.

\emph{Manual rigs.} As an assist, the same filter could show the predicted swing to the driller or drive a powered guide, so that a floorhand does not have to steady the joint by hand.

\emph{Other suspended loads.} Any job that brings a crane- or hoist-hung load to a fixed target has the same structure: steel erection \citep{jung2007gripping}, offshore pile and monopile stabbing, wind-turbine tower and blade assembly, and pipe handling on vessels. The estimator, the matched-speed capture and the touch-confirmed approach carry over, with the head and the tuning adapted to the load.

\section{Scope and Next Steps}
The study is a full-scale-equivalent simulation of capture, centring and stab-in; make-up, real images and rig-specific effects such as the elevator links are left to the next stage. That stage is a test stand with a real stereo camera and instrumented pipe, fine-tuning of the network on labelled real images, and a force-based check of stab quality before make-up \citep{park2018multimodal,ruehmann2021autonomous}.

\section{Conclusion}
\begin{itemize}
\item A robot that predicts the swing of a hanging joint and meets it at its own speed seated all 200 random jobs across nine sizes, from 2$\frac{3}{8}$-in. tubing to 20-in. casing, with the pipe touching nothing but rollers; a fixed guide on the same jobs seated 74\%.
\item Median contact speed and median peak force both fell by a factor of 7.8, the 95th-percentile force by a factor of 9.9, and the impact energy at first contact by a median factor of 95; the force was lower in every job.
\item StabBot removed the missed captures and steel hits that ended one fixed-guide job in four, and takes the floorhand's hand off the pipe during the stab.
\item Each part adds measurable value: the guarded approach made the capture reliable, prediction with velocity matching made it gentle, and residual learning lowered peak force by a further 22\%.
\item With the stereo keypoint network in the loop, success stayed at 100\% under camera delay, frame loss, calibration error, frequency error, an off-centre elevator, swings of up to 40\,cm and up to 20\,mm of added measurement noise.
\item The network located the pipe, the pin and the box with sub-millimetre median error in 2.3\,ms per stereo pair.
\item One open V head on free rollers served all nine sizes, and the method carries over to other hoist-hung loads brought to a fixed target.
\end{itemize}

\Needspace{9\baselineskip}
\section*{Abbreviations}
\begin{multicols}{2}\noindent
\begin{tabular}{@{\hspace{2em}}p{0.16\columnwidth}l@{}}
3-D & Three-dimensional\\
CI & Confidence interval\\
CNN & Convolutional neural network\\
F/T & Force--torque (sensor)\\
GPU & Graphics processing unit\\
MuJoCo & Multi-Joint dynamics with Contact\\
\end{tabular}
\columnbreak
\noindent
\begin{tabular}{@{\hspace{2em}}p{0.16\columnwidth}l@{}}
MLP & Multilayer perceptron\\
p95 & 95th percentile\\
PPO & Proximal policy optimization\\
ResNet & Residual network\\
RL & Reinforcement learning\\
\end{tabular}
\end{multicols}

\Needspace{6\baselineskip}
\section*{Nomenclature}
\begin{longtable}{@{\hspace{2em}}p{0.13\linewidth}p{0.62\linewidth}l@{}}
$c$ & Rest position about which the joint swings & m\\
$F_k$, $F_{\max}$ & Peak head--pipe force in a decision step, and over the episode & kN\\
$k$ & Filter step or decision index & --\\
$\mathbf n$ & Unit vector along the approach direction & --\\
$o$ & Offset of the elevator from well centre & m\\
$p$, $v$ & Position and velocity of the pipe axis at head height & m, m/s\\
$\hat{\mathbf p}$ & Predicted pipe position & m\\
$P_m$ & Share of jobs in which the fixed guide missed the capture & --\\
$r$ & Ratio of the head-to-pin and elevator-to-head heights & --\\
$r_k$ & Reward for decision $k$ & --\\
$s$ & Stand-off along the approach direction & m\\
$t$ & Time & s\\
$T_G$, $T_S$ & Median time to seat, fixed guide and StabBot & s\\
$\bar T_G$, $\bar T_S$ & Expected time per connection, fixed guide and StabBot & s\\
$T_r$, $T_r^{*}$ & Re-stab time, and the value above which StabBot is faster & s\\
$u_c$ & Contact speed at the first roller contact & cm/s\\
$\mathbf x$ & Kalman filter state $[p,\,v,\,c]^\top$ & m, m/s, m\\
$x_h$ & Hold position of the pipe at head height & m\\
$\mathbf y^*$ & Target position of the V seat & m\\
$z_e$, $z_h$, $z_p$ & Heights of the elevator, the head and the pin nose & m\\
$\Delta t$ & Time between filter steps & s\\
$\delta^{c}$, $\delta^{s}$, $\delta^{f}$, $\delta^{e}$ & Event indicators: first contact, centred, failure, last step & --\\
$\gamma$, $\lambda$ & PPO discount factor and advantage-estimation parameter & --\\
$\sigma$ & Width of the Gaussian heatmap target & px\\
$\tau$ & Prediction look-ahead & s\\
$\omega$ & Angular frequency of the swing & rad/s\\
\end{longtable}

\section*{Statements and Declarations}
\noindent\textbf{Funding.} No funding was received to assist with the preparation of this manuscript.

\noindent\textbf{Competing interests.} The authors declare that they have no known competing financial interests or personal relationships that could have appeared to influence the work reported in this paper.

\noindent\textbf{Ethics approval and consent to participate.} Not applicable. The study is a computer simulation; it involves no human participants, no animal subjects and no personal data.

\noindent\textbf{Consent for publication.} Not applicable: the article contains no data relating to an identifiable individual.

\noindent\textbf{Data availability.} The recorded benchmark jobs, the robustness and unseen-condition results and the files behind every table and figure are kept in one repository and will be made public on acceptance. They are available to reviewers on request.

\noindent\textbf{Code availability.} The simulation, controllers, training and analysis scripts, the trained keypoint networks (v2, v3) and the residual policy will be released with the data; the corresponding author will share them with reviewers on request.

\noindent\textbf{Author contributions.} Mohamed Ashraf Mostafa conceived the study, designed the methodology, wrote the software, ran the investigation and the formal analysis, made the visualisations and wrote the original draft. Taher El-Fakharany and Samir Khaled supervised the work and both reviewed and edited the manuscript.

\section*{Declaration of generative AI and AI-assisted technologies in the manuscript preparation process}
This research is the authors' own work. During its preparation the authors used Claude (Anthropic) as an assistant, directed by the first author, for help with coding, drafting and language review. The authors reviewed and edited all output and take full responsibility for the publication. All results, numbers and figures come from the authors' own simulations.

{\footnotesize
\setlength{\bibsep}{3pt plus 0.5pt}
\bibliographystyle{elsarticle-num}
\bibliography{refs}
}

\end{document}